\documentclass[letterpaper, 10 pt, journal, twoside]{IEEETran}

\usepackage[english]{babel}
\usepackage{amsmath}
\usepackage{amssymb}
\usepackage{booktabs}
\usepackage{caption}
\usepackage{subfigure}
\usepackage[dvips]{graphicx}
\usepackage{multirow}
\usepackage{amsfonts}
\usepackage{breakurl}
\usepackage{enumerate}
\usepackage{tabularx}
\usepackage{algorithm,algorithmic}
\usepackage{bm}
\usepackage{enumerate}
\usepackage{adjustbox}
\usepackage{xcolor}
\usepackage{siunitx}
\usepackage{booktabs}
\usepackage{mdframed}
\usepackage{adjustbox}
\usepackage{authblk}
\usepackage{color}
\usepackage{xurl}
\usepackage{cite}
\makeatletter
\let\NAT@parse\undefined
\makeatother
\usepackage{hyperref}
\newcolumntype{s}{>{\hsize=.3\hsize}X}
\newcolumntype{x}{>{\hsize=.5\hsize}X}
\usepackage{tikz}
\usetikzlibrary{shapes.geometric, arrows}
\usepackage{svg}

\tikzstyle{startstop} = [rectangle, rounded corners, minimum width=3cm, minimum height=1cm,text centered, draw=black, fill=red!30]
\tikzstyle{arrow} = [thick,->,>=stealth]

\title{\LARGE \bf From Propeller Damage Estimation and Adaptation \\
to Fault Tolerant Control: Enhancing Quadrotor Resilience}

\author{Jeffrey Mao*, Jennifer Yeom*, Suraj Nair, and Giuseppe Loianno
\thanks{$^*$These authors contributed equally and are listed in alphabetic order.}
\thanks{Manuscript received: October 19, 2023; Accepted February 28, 2024.}
\thanks{This paper was recommended for publication by Editor  Cesar Cadena and Editor Lucia Pallottino upon evaluation of reviewers' comments.
This work was supported by the DEVCOM ARL grant DCIST CRA W911NF-17-2-0181, NSF CAREER Award 2145277, DARPA YFA Grant D22AP00156-00, Qualcomm Research, Nokia, and NYU Wireless.} 
\thanks{The authors are with the New York University, Tandon School of Engineering, Brooklyn, NY 11201, USA. {\tt\footnotesize email: \{jm7752, jennifer.yeom, surajkiron, loiannog\}@nyu.edu}. (Jeffrey Mao
and Jennifer Yeom are co-first authors.) (Corresponding author: Jeffrey Mao.)}
\thanks{Digital Object Identifier (DOI): see top of this page}
}
\begin{document}
\maketitle

\begin{abstract}
Aerial robots are required to remain operational even in the event of system disturbances, damages, or failures to ensure resilient and robust task completion and safety. One common failure case is propeller damage, which presents a significant challenge in both quantification and compensation. In this paper, we propose a novel adaptive control scheme capable of detecting and compensating for multi-rotor propeller damages, ensuring safe and robust flight performances. Our solution combines an L1 adaptive controller  with an optimization routine for damage inference and compensation of single or dual propellers, with the capability to seamlessly transition to a fault-tolerant solution in case the damage becomes severe. We experimentally identify the conditions under which the L1 adaptive solution remains preferable over a fault-tolerant alternative. Experimental results validate the proposed approach demonstrating the ability of our solution to adapt and compensate onboard in real time on a quadrotor for damages even when multiple propellers are damaged.
\end{abstract}
\begin{IEEEkeywords}
Aerial Systems: Mechanics and Control; Aerial Systems: Applications
\end{IEEEkeywords}

\section*{Supplementary material}
\textbf{Video}: \url{https://youtu.be/elVO0-tkPs0}

\IEEEpeerreviewmaketitle

\section{Introduction} \label{sec:introduction}
\IEEEPARstart{M}{icro} Aerial Vehicles (MAVs) such as quadrotors are becoming ubiquitous in applications such as search and rescue and aerial photography~\cite{idrissi2022review}. However, MAV-related accidents hinder the growth of the industry and erode public confidence in drone safety. As a result, it is necessary to develop inference and control approaches that can guarantee safe and reliable flight in case of system damage.
Propellers on MAVs are susceptible to damage, especially in the event of collisions or after prolonged use, primarily due to their size, location, and lightweight construction. In addition, detecting the occurrence and location of propeller failures is challenging. Indirectly sensing individual motor thrusts is further complicated due to the nonlinear dynamics of the quadrotor and the difficulty of measuring higher order angular acceleration or moment terms.
Our work specifically introduces an adaptive control scheme for quadrotors to autonomously infer and adjust flight behavior in response to propeller damages or failures. 

\begin{figure}
    \centering
    \includegraphics[width= \linewidth]{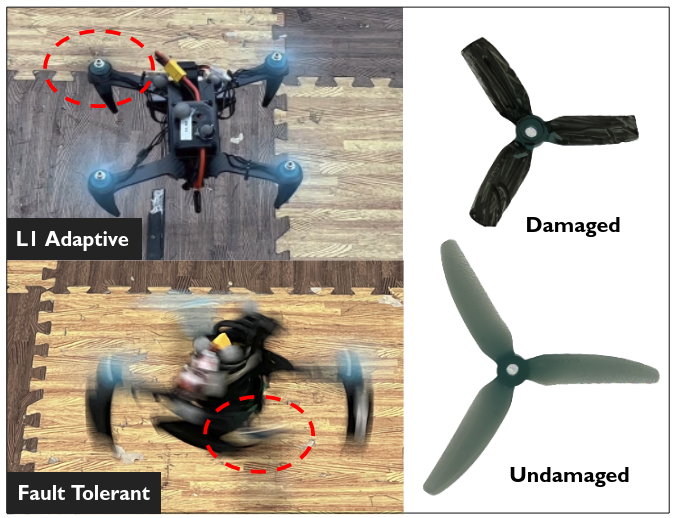}
    \caption{L1 adaptive and fault-tolerant control with one damaged propeller (circled in red dash). The fault-tolerant controlled quadrotor is forced to spin (over $1000$ deg/sec).}
    \vspace{-0.6cm}
    \label{fig:intro_image}
\end{figure}

This work makes the following contributions. 
First, we propose a propeller damage estimation technique by combining an L1 Adaptive algorithm and an optimization routine with no need for direct RPM (Rotations Per Minute) measurements or estimation of angular accelerations, making it suitable for low-cost systems. 
Second, we propose a holistic control scheme coupling the L1 adaptive and fault-tolerant controller modules for autonomous transition from L1 adaptation to the fault-tolerant mode in situations where adaptation alone proves to be insufficient for ensuring safe and effective flight. 
This approach enhances the system's capability to manage the full range of propeller damage.
Finally, we demonstrate the performance of the damage estimation and controller design in extensive real-world experiments and compare our estimation method to using a standard PID (Proportional Integral Derivative) controller.
This is the first work to tackle precise estimation and compensation of propeller damage 
rather than considering a reduction of effectiveness of a motor-propeller pair. 
Challenges include unequal loss between thrust and torque coefficients along with additional noise on the system from damaged propellers as well as the unavailability of direct RPM measurements. 

\section{Related Works}~\label{sec:related_works}
\textbf{Damage Estimation}. Research in fault estimation covers a variety of vehicles detailed in this survey \cite{fourlas2021survey} such as fixed wings and rotary aerial robots. We can divide methods handling propeller damage into two categories: empirical and state estimation methods. Empirical works address propeller damage estimation through parameter identification \cite{peng_fault_detection} or sensor noise analysis \cite{vibration_mark}. 
The approach proposed in~\cite{peng_fault_detection} successfully estimates the loss of effectiveness of a single motor but is only tested in simulation, whereas~\cite{vibration_mark} uses accelerometer noise to estimate propeller damage, but is only tested on a minuscule amount of propeller damage. 

Other works rely on state estimation to accurately measure the damage \cite{EKF_adapt_est_motor_failure, Avram_2017_detect_adapt}. These typically have a slower response time than empirical methods, and are only able to handle motor damage estimation rather than propeller damage as in the proposed work.
They typically assume the damage from the propeller motor pair can be represented as a scaled factor of each individual RPM. For the case of propeller damage, this assumption is not accurate. The thrust and torque coefficients scale with the propeller radius, $r$, as a function of $r^4$ and $r^5$ respectively \cite{quadrotor_modelling}. This increases the complexity of our propeller damage compared to the problems in \cite{EKF_adapt_est_motor_failure, Avram_2017_detect_adapt, schijndel_2023}.
Typically a reduced state Kalman Filter \cite{alex2016reduced, EKF_adapt_est_motor_failure, Avram_2017_detect_adapt} is used in either a single \cite{alex2016reduced, EKF_adapt_est_motor_failure} or cascaded\cite{ROT2020682, zhong_robust} form.
These studies test their systems by artificially injecting a fault that decelerates the motor rather than testing true propeller damages. State estimation through cascaded Kalman Filters can infer and adapt up to $60\%$ motor damage in~\cite{zhong_robust}. However, it is only tested in simulation for motor damage and does not include the possibility to switch to the fault-tolerant mode. Our work negates up to $40-60\%$ propeller damage in real-world experiments as shown in Fig.~\ref{fig:intro_image}. 
Uniquely, \cite{schijndel_2023} implements a Kalman filter method to detect a complete loss of a propeller and is able to detect a fault within $0.1~\si{s}$ of occurrence. However, this method requires accurate RPM measurements and cannot adapt to propeller damage. 
 
\textbf{Damage Compensation}. Once propeller damage is estimated, there are two options to incorporate damage compensation into control. The first option is adaptation where the damaged propeller is given a scaled up motor action \cite{EKF_adapt_est_motor_failure, Avram_2017_detect_adapt} to compensate for the damage. 
This adaptation is effective for small and medium damages but is unable to stabilize the robot for severe damages.
The second option is to employ a fault-tolerant control strategy that completely disables the damaged propeller \cite{peng_fault_detection, Abbaspour2020ASO} but sacrifices yaw control. 

A few adaptive techniques are Incremental Nonlinear Dynamic Inversion (INDI)~\cite{sun2021indi}, dual loop disturbance observers~\cite{xiang_sfc}, and L1~\cite{NairaL1, muhlegg_2014, min_2018}. These works provide additional control actions that augment the capabilities of a baseline controller to reject some actuator damages or disturbances.
However, none of the above works provides an estimate to judge when propeller damage is too severe for adaption as shown in the proposed work. 
Other methods exist for adaptation~\cite{saviolo2022active, crocetti2023gapt, loquercio2022autotune}, through model identification or automatic gain tuning, but cannot adapt to rapid changes. 
In this work, we design an L1 adaptive technique to compensate and accurately estimate  propeller damage.
Unlike INDI\cite{sun2021indi} that requires angular acceleration which is difficult to estimate, our method only requires angular velocity which can be obtained from an on board Inertial Measurement Unit (IMU).

Existing fault-tolerant control strategies are adept at ensuring robust control even in the event of single or multiple rotor failures~\cite{yeom2023geometric,sun2021indi, fourlas2021survey, mueller2014lqr}.
However, these solutions require identifying the faulty propeller, and sacrificing yaw control which results in a rapid yaw spin as depicted in Fig.~\ref{fig:intro_image} making them unsuitable for small to medium damages. 
Our work introduces a holistic control scheme designed to detect, infer, and compensate for propeller damage, with the capability to smoothly transition to a fault-tolerant solution when severe damage is detected allowing effective operations for safe recovery and navigation.

\begin{figure}
    \centering
       \includegraphics[width= \linewidth]{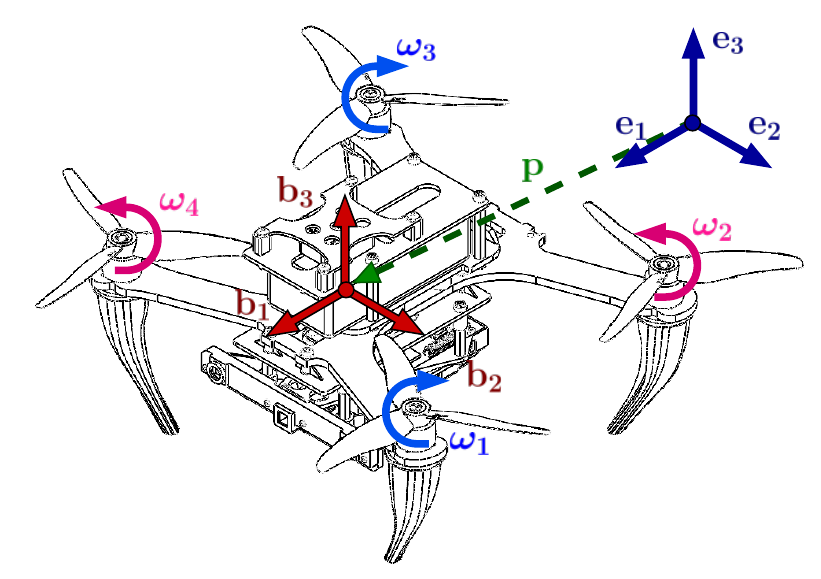}
    \vspace{-2.0em}
    \caption{MAV model with inertial and body frame definitions.}
    \label{fig:quadrotor_model}
    \vspace{-1.8em}
\end{figure}

\section{Methodology} \label{sec:methodology}
\subsection{Preliminaries}
We use two coordinate frames to represent the dynamics of the system. 
As shown in Fig.~\ref{fig:quadrotor_model}, we choose the inertial reference frame as $\begin{bmatrix}\mathbf{e}_1& \mathbf{e}_2& \mathbf{e}_3\end{bmatrix}$ and the body fixed frame as $\begin{bmatrix}\mathbf{b}_1& \mathbf{b}_2& \mathbf{b}_3\end{bmatrix}$.
The origin of the body frame is aligned with the center of mass of the MAV. The body frame follows the East North Up (ENU) coordinate system. The first axis $\mathbf{b}_1$ is aligned to the heading and the third axis, $\mathbf{b}_3$  aligns with the thrust vector of the vehicle. 
The dynamics of the system can be written as
\begin{equation} \label{eq:1}
\begin{split}
    \dot{\mathbf{p}} &= \mathbf{v},\\
    m \dot{\mathbf{v}} &= m g \mathbf{e}_3 +  \mathbf{q} \otimes f\mathbf{e}_3 ,\\
    \dot{\mathbf{q}} &= \frac{1}{2}\mathbf{q} \otimes  \begin{bmatrix}
        0 \\
        \mathbf{\Omega} \\
    \end{bmatrix},\\
    \mathbf{M} &= \mathbf{J} \dot{\mathbf{\Omega}} + \mathbf{\Omega} \times \mathbf{J}\mathbf{\Omega} ,
\end{split}
\end{equation}
where $\mathbf{p} \in \mathbb{R}^3$, and $\mathbf{v}\in \mathbb{R}^3$ are the position and velocity in the world frame. $\mathbf{R}\in SO(3)$ is a rotation matrix from the body fixed frame to the inertial frame, $\mathbf{q}$ is the quaternion representation of $\mathbf{R}$,  and $\otimes$ is quaternion multiplication. $f$, $m$, $g$ are the thrust, mass and gravity respectively, $\mathbf{\Omega}\in \mathbb{R}^3$ is the angular velocity with respect to the body frame, $\mathbf{M}=\begin{bmatrix} M_1& M_2& M_3\end{bmatrix}^\top$ are the moments around the three body frame axes, and $\mathbf{J}\in \mathbb{R}^{3 \times 3}$ is the robot inertia matrix. 

\begin{figure}[!t]
    \centering
    \includegraphics[width= 0.99\linewidth]{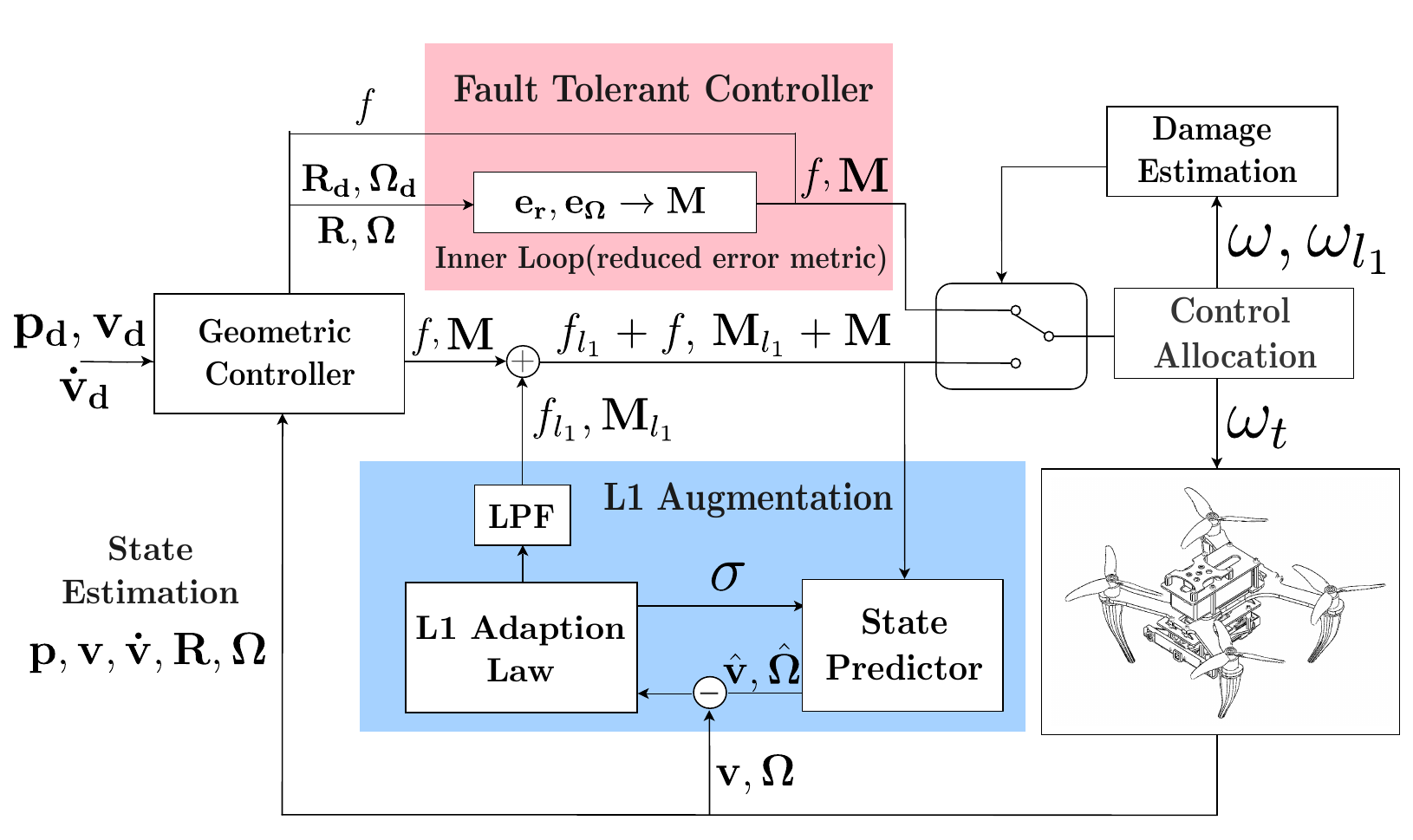}
    \caption{The cascaded control scheme, shows the L1 Adaptation Controller (blue), receives the partial state feedback from the system. The Fault Tolerant Controller (pink) is activated only once damage estimate exceeds $50\%$.}
    \vspace{-18pt}
    \label{fig:control_diag}
\end{figure}
\subsection{Control Design} \label{sec:control_design}
We adapt the geometric controller derived in \cite{lee2010so3} for the possibility of propeller damage, or fully losing one or two opposing rotors in flight. An outer controller solves for the desired attitude $\mathbf{R}$, thrust $f$, and angular velocities $\mathbf{\Omega}$. The inner controller solves for the moments required to control the attitude of the MAV. 
The geometric controller tracks position $\mathbf{p}$ and yaw $\psi$ by generating force and moments control signals.
The L1 augmentation is inserted into the controller as seen in Fig.~\ref{fig:control_diag}. The L1 controller uses a state predictor to estimate the partial state $[\mathbf{v}^\top~ \mathbf{\Omega}^\top]^\top$ with the unknown disturbances. The adaptation law compares the prediction to the measured state and generates the control output to compensate for disturbances. If the adaptation law accurately estimates the unknown disturbances then the tracking error will converge to zero \cite{stab_l1} assuming the augmented action is with in the actuator limits. 
\subsubsection{Outer-loop Controller}
As per the standard geometric control algorithm, the thrust vector is normalized and chosen as the third body axis, or thrust vector $\mathbf{b}_3$ as

\begin{equation}\label{eq:b3}
    \mathbf{b}_3 = \frac{\dot{\mathbf{v}}+g\mathbf{e}_3}{\|\dot{\mathbf{v}}+g\mathbf{e}_3 \|}.
\end{equation}
Next, we formulate the rotation matrix based on $\psi$ or the yaw of the vehicle. We denote the individual elements of the thrust vector as $\mathbf{b}_3 = \begin{bmatrix}b_{3x} & b_{3y} & b_{3z} \end{bmatrix}^\top$. We can formulate a quaternion representing the desired tilt, $\mathbf{q}_{tilt}$, and yaw rotation $\mathbf{q}_{\psi}$ such as in \cite{watterson2019control}. The below quaternion formulations are all scalar first notation.
\begin{equation} \label{eq:rot_comp}\begin{split}
\mathbf{q}_{tilt} &= \frac{1}{\sqrt{2(1+b_{3z})}}\begin{bmatrix}
    1+b_{3z} & -b_{3y} & b_{3x}& 0\\
\end{bmatrix}, \\
\mathbf{q}_\psi &= \begin{bmatrix}
    \cos(0.5\psi)  & 0 & 0& \sin(0.5\psi)\\
\end{bmatrix}.
\end{split}
\end{equation}
Multiplying the quaternion  $\mathbf{q}_{tilt}$ and $\mathbf{q}_{\psi}$ we formulate the full vehicle's rotation with $\mathbf{q}_d$ being the quaternion representation of desired orientation, $\mathbf{R}_d$. A rotation matrices representation of eq.~(\ref{eq:rot_comp}) is in \cite{yeom2023geometric}
\begin{equation} 
\label{eq:rot}
\mathbf{q}_d = \mathbf{q}_{tilt} \otimes \mathbf{q}_\psi.
\end{equation}
The first control input, thrust, is solved for with
\begin{equation} 
f = m \left(\mathbf{k}_p ( \mathbf{p} - \mathbf{p}_d) + \mathbf{k}_v (\mathbf{v} - \mathbf{v}_d)+  \dot{\mathbf{v}}_d + g \mathbf{e}_3\right) \cdot \mathbf{b}_3,
\label{eq:thrust_command}
\end{equation}
where $\mathbf{p}_d$ and $\mathbf{v}_d$ are desired position and velocity of the vehicle and $\mathbf{k}_p, \mathbf{k}_v \in\mathbb{R}^3$ are the gains for the respective errors. 
Lastly, the desired angular velocity $\mathbf{\Omega}_d$, is 
\begin{equation} \label{eq:omega_hat}
    \begin{bmatrix}
        0 \\
        \mathbf{\Omega}_d \\
    \end{bmatrix} = 2 \mathbf{q}_d^{-1} \otimes \dot{\mathbf{q}_d},
\end{equation} 
as a function of the desired $\dot{\mathbf{b}}_3$ and $\dot{\psi}$ such as in \cite{watterson2019control} .

\subsubsection{Inner-loop Controller}
The inner controller solves for the second control input, moment, from the error in attitude and error in angular velocity. In a traditional geometric controller \cite{lee2010so3}, the attitude tracking error, $\mathbf{e}_R$ is calculated by using the rotation matrices as
\begin{equation} \label{eq:e_R}
    \mathbf{e}_R = \frac{1}{2} (\mathbf{R}_d^\top \mathbf{R} - \mathbf{R}^\top \mathbf{R}_d)^{\vee},
\end{equation}
where the $^{\vee}$ term converts a skew-symmetric matrix to vector form. For the fault-tolerant controller, we use a reduced attitude error metric to decouple the yaw control that shows improved performances as demonstrated in our recenter work~\cite{yeom2023geometric} for the fault-tolerant control case
\begin{equation} \label{eq:e_thrust_vector}
    \mathbf{e}_{R_{\text{reduced}}} = \mathbf{b}_{3d} \times \mathbf{b}_3,
\end{equation}
where $\mathbf{b}_{3d}$ is the normalized desired thrust vector and $\mathbf{b}_3$ is the current $z$ axis of the body fixed frame we get from the localization. The cross product of $\mathbf{b}_{3d}$ and $\mathbf{b}_3$ measures how much the actual direction deviates from the desired direction.
The angular velocity error, $\mathbf{e}_\Omega$ is calculated by 
\begin{equation} \label{eq:e_omega}
\mathbf{e}_\Omega = \mathbf{\Omega} - \mathbf{R}^\top \mathbf{R}_d \mathbf{\Omega}_d.
\end{equation}
The moments are calculated using
\begin{equation}
\mathbf{M} = -\mathbf{k}_R \mathbf{e}_R - \mathbf{k}_{\Omega} \mathbf{e}_{\Omega} + \mathbf{\Omega}\times \mathbf{J} \mathbf{\Omega},
\label{eq:moment_commands}
\end{equation}
where $\mathbf{k}_R$ and $\mathbf{k}_{\Omega}$ are rotation and angular velocity gains.

\subsection{Adaptive Control}
We supplement our controller with an L1 adaptive controller by including two elements:
$1)$ an adaption law to solve for an action to counteract the un-modeled disturbances on the system, and 
$2)$ a state predictor to estimate a predicted state, linear $\mathbf{v}_p$ and angular $\mathbf{\Omega}_p$ velocities  given the L1 action and un-modeled disturbances.

\subsubsection{L1 Adaptation Law}
\begin{algorithm}[b]
\caption{L1 Adaptation Control Step $[k] \rightarrow [k+1]$ \label{alg_l1} }
\begin{algorithmic}[1]
    \item[\textbf{Input:}]  Measured velocities  $\mathbf{v}_m[k], \mathbf{\Omega}_m[k]$
    \item[\textbf{Output:}] L1 Force and Moment $f_{L1}[k], \mathbf{M}_{L1}[k]$
    \item[\textbf{State:}] Predicted velocities $\mathbf{v}_p[k], \mathbf{\Omega}_p[k]$ \\
    \item[\textbf{L1 Adaption control law Section~\ref{sec:methodology}.C.1}]  
  \STATE Calculate the Disturbances, $\bm{\sigma}[k]$ (eq.~(\ref{eq:gain}))
  \STATE $f_{L1}[k] = \text{LPF}(-\bm{\sigma}_3[k])$  \STATE $\mathbf{M}_{L1}[k] = \text{LPF}(-\bm{\sigma}_{4:6}[k])$
    \item[\textbf{State Predictor (Update) Section~\ref{sec:methodology}.C.2}]  
  \STATE Solve for $\dot{\mathbf{v}}_p[k]$ (eq.~(\ref{eq:vel_state})) and $\dot{\mathbf{\Omega}}_p[k]$ (eq.~(\ref{eq:angular_state}))  
  \STATE Solve for $\mathbf{v}_p[k+1]$ and $\mathbf{\Omega}_p[k+1]$ (eq.~(\ref{eq:state_update}))
  \STATE Update the Current $[k]$ with the Next State $[k+1]$
  \STATE Add $f_{L1}[k]$ the base thrust control eq.~(\ref{eq:thrust_command}) 
  \STATE Add $\mathbf{M}_{L1}[k]$ to the base moment control eq.~(\ref{eq:moment_commands})      
\end{algorithmic}
\end{algorithm}

First, we characterize the disturbances $\bm{\sigma}=\begin{bmatrix}\mathbf{f}_{\text{ext}}^\top &\mathbf{M}_{\text{ext}}^\top\end{bmatrix}^\top$ which represents the  un-modeled force and moments on the vehicle both expressed in the body frame. These disturbances are obtained using the error between the predicted state velocity $\begin{bmatrix}\mathbf{v}_p^\top & \mathbf{\Omega}_p^\top\end{bmatrix}$ generated from the state predictor in Section~\ref{sec:methodology}.C.2 and measured velocities, $(\mathbf{v}_m, \mathbf{\Omega}_m)$. Let $k$ denote the discrete time step of the system.  
We set a diagonal gain defined by the user $\mathbf{K} \in \mathbb{R}^{6\times6}$ to generate a specific gain, $\mathbf{A}$, based on the time step, $dt$, between $k$ and $k+1$ time instants with 
\begin{equation}
\label{eq:gain}
\begin{split}
    \mathbf{A} &= (\exp(\mathbf{K} \cdot dt-\mathbf{I})^{-1}\mathbf{K})\exp(\mathbf{K} \cdot dt),\\ 
\bm{\sigma}[k] &= \begin{bmatrix}
    m \mathbf{R}^\top & 0 \\
    0 & \mathbf{J} \\
\end{bmatrix}\mathbf{A}\left(\begin{bmatrix}
    \mathbf{v}_m[k] \\ \mathbf{\Omega}_m[k]\end{bmatrix}-\begin{bmatrix}
    \mathbf{v}_p[k] \\ \mathbf{\Omega}_p[k] \end{bmatrix}\right).
\end{split}
\end{equation}
 We solve the additional action $\bm{\mu}_{L1} = \begin{bmatrix}f_{L1} & \mathbf{M}_{L1}^\top\end{bmatrix}^\top$ from our disturbance $\bm{\sigma}$.  The moment disturbance estimates $\bm{\sigma}_{4:6}$ and disturbance along the thrust axis $\bm{\sigma}_{3}$ can be negated by providing an equal but negative compensation to cancel the disturbance $\bm{\mu}_{L1} = -\bm{\sigma}_{3:6}$.  We implement an exponential Low Pass Filter (LPF) to mitigate the noise present in the state estimation, achieving smoother results: $\bm{\mu}_{L1} =~-\text{LPF}(\bm{\sigma}_{3:6})$. $f_{L1}$ and $\mathbf{M}_{L1}$ are added to eq.~(\ref{eq:thrust_command}) and eq.~(\ref{eq:moment_commands}) respectively and reflected in Fig.~(\ref{fig:control_diag}).

\subsubsection{State Predictor}
Separately, once we solve for the action $\bm{\mu}_{L1}$ and external disturbances $\bm{\sigma}$, we solve for the next predicted velocities of our system, $(\mathbf{v}_p[k], \mathbf{\Omega}_p[k]) \rightarrow (\mathbf{v}_p[k+1], \mathbf{\Omega}_p[k+1])$. 
  First, we solve for the linear and angular acceleration $\dot{\mathbf{v}}_p[k]$ and $\dot{\mathbf{\Omega}}_p[k]$ as 
\begin{equation} \label{eq:vel_state}
\mathbf{\dot{v}}_p[k] =  \underbrace{\frac{(f[k]+f_{L1}[k])}{m} \mathbf{R} \mathbf{e}_3}_{\text{actuation}}  +\underbrace{\frac{1}{m}\mathbf{R} \bm{\sigma}_{1:3}[k]}_{\text{disturbance}} +\underbrace{g \mathbf{e}_3}_{\text{gravity}},
\end{equation}
\begin{equation} \label{eq:angular_state}
\mathbf{\dot{\Omega}}_p[k] = \mathbf{J}^{-1}(\underbrace{\mathbf{M}[k]+\mathbf{M}_{L1}[k]}_{\text{actuation}} + \underbrace{\bm{\sigma}_{4:6}[k]}_{\text{disturbance}}-\underbrace{\mathbf{\Omega}_m[k] \times \mathbf{J}\mathbf{\Omega}_m[k]}_{\text{cross}} ) .
\end{equation}
Finally, we integrate over a time step $dt$ with a weighted average of the measurement and the predicted velocities. A weighted average is used to filter the noise inherent in the measured velocity 
\begin{equation} \label{eq:state_update}
    \begin{bmatrix}
    \mathbf{v}_p[k+1] \\ \mathbf{\Omega}_p[k+1] \end{bmatrix} =  (1-\lambda)\begin{bmatrix}
     \mathbf{v}_p[k] \\  \mathbf{\Omega}_p[k]  \end{bmatrix}  + \lambda\begin{bmatrix}
     \mathbf{v}_m[k] \\  \mathbf{\Omega}_m[k]  \end{bmatrix}  + dt \begin{bmatrix} 
    \dot{\mathbf{v}}_p[k] \\ \dot{\mathbf{\Omega}}_p[k] \end{bmatrix}.
\end{equation}
The weighted average term is $\lambda = \mathbf{K}\cdot dt $. This is predicted value is looped backed into our L1 control law eq.~(\ref{eq:gain}) in the next time step $k+1$

\begin{table*}[!t]
\centering
\caption {Tracking RMSE (in meters) where $k_{f_{\text{mis}}}$ represents the thrust loss. FT represents the fault-tolerant control. L1 Off does not have data for $k_{f_{mis}}=60\%$ because the aerial robot is unstable outside of the hover condition.}
\resizebox{\textwidth}{!}{
\begin{tabular}{c c c c c c c c c c c  }
\toprule\toprule
 \rule{0pt}{2ex} Duration $[\si{s}]$ & Axis & \multicolumn{2}{c}{$k_{f_{\text{mis}}} = 0.0$} & \multicolumn{2}{c}{$k_{f_{\text{mis}}} = 20\%$}&\multicolumn{2}{c}{$k_{f_{\text{mis}}} = 40\%$}&\multicolumn{2}{c}{$k_{f_{\text{mis}}} = 60\%$} & {FT}\vspace{2pt}
\\
  \cline{3-11}\
  \rule{0pt}{2ex} Max Speed $[\si{m/s}]$ & &  L1 ON    & L1 OFF  &  L1 ON   & L1 OFF &  L1 ON    & L1 OFF&  L1 ON    & L1 OFF&  L1 ON\\
\hline
\multirow{3}{*}{\shortstack{12 \\$[0.5]$} }
& $x$  &  $\mathbf{0.031}$  &  0.042  &  $\mathbf{0.017}$ & 0.072  &  $\mathbf{0.020}$ & 0.127  & $\mathbf{0.207}$ &  -  &  0.041  \\
& $y$  &  $\mathbf{0.070}$   &  0.092  &  $\mathbf{0.062}$ & 0.233  &  $\mathbf{0.030}$ & 0.453 & $\mathbf{0.358}$ &  -  &  0.044  \\
& $z$  &  $\mathbf{0.002}$  &  0.017  &  $\mathbf{0.002}$ & 0.050  &  $\mathbf{0.003}$ & 0.082  & $\mathbf{0.005}$ &  -  &  0.011  \\
\hline

\multirow{3}{*}{\shortstack{8\\$[0.8]$}}
& $x$  &  $\mathbf{0.037}$  &  0.040  &  $\mathbf{0.031}$ & 0.067   &  $\mathbf{0.032}$ & 0.16  & $\mathbf{0.208}$ &  -  &  0.075  \\
& $y$  &  $\mathbf{0.064}$  &  0.10  &  $\mathbf{0.064}$ & 0.222   &  $\mathbf{0.048}$ & 0.42  & $\mathbf{0.384}$ &  -  &  0.059  \\
& $z$  &  $\mathbf{0.002}$  &  0.022  &  $\mathbf{0.004}$ & 0.051   &  $\mathbf{0.003}$ & 0.09  & $\mathbf{0.007}$ &  -  &  0.020  \\
\hline
\multirow{3}{*}{\shortstack{5 \\$[1.1]$}}
& $x$   &  $\mathbf{0.045}$  &  0.052  &  $\mathbf{0.060}$ & 0.079   &  $\mathbf{0.046}$ & 0.181  & $\mathbf{0.204}$ &  -  &  0.093  \\
& $y$  &   $\mathbf{0.079}$  &  0.112  &  $\mathbf{0.072}$ & 0.249   &  $\mathbf{0.047}$ & 0.406 & $\mathbf{0.406}$ &  -  &  0.109  \\
& $z$   &  $\mathbf{0.003}$  &  0.025  &  $\mathbf{0.005}$ & 0.050   &  $\mathbf{0.003}$ & 0.090  & $\mathbf{0.009}$ &  -  &  0.019  \\
\hline\hline\end{tabular}
}
\label{tab:rmse1}
\vspace{-1.3em}
\end{table*}

\subsection{Propeller Damage Estimation}~\label{sec:estpropeller_damage}
We estimate the propellers' damage using our augmented control. 
We assume that in nominal flight conditions a quadratic relationship between the $i^\text{th}$ propeller thrust and corresponding motor speed $f_i= k_{f_{\text{real}}}\omega_{i}^2$ where $k_{f_{\text{real}}}$ represents the true thrust coefficient of the propeller and $\omega_{i}$ is the motor speed of the $i^\text{th}$ propeller. 
A damaged propeller spinning at the original RPM provides lower thrust due to the change in $k_{f_{\text{real}}}$. We quantify our propeller damages on the thrust coefficient by the following mismatch index
\begin{equation}\label{eqn:prop_dam}
    k_{f_{\text{mis}}} = 1-\frac{k_{f_{\text{real}}}}{k_{f_{\text{model}}}},
\end{equation}
where $k_{f_{\text{model}}}$ is the thrust coefficient used by our controller. 
When a propeller gets damaged, both the thrust coefficient $k_f$ and torque coefficient $k_m$ are affected. The two coefficients decay as functions of the propeller radius $r^4$ and $r^5$ respectively \cite{quadrotor_modelling}.
Due to this relationship, a damaged propeller spinning at higher RPM will provide adequate thrust but less torque than a healthy propeller. 
The L1 adaptive law compensates for both thrust and torque mismatch requiring the healthy propellers' RPMs to compensate for the damaged propeller's lower torque.
As the thrust and torque coefficients are highly coupled, we formulate an optimization problem for damage estimation.
First, we initialize a baseline estimate of our propeller damage for each propeller.
Next, we identify damaged and undamaged propellers to form a prior.
Finally, we perform a null-space projection optimization.

We can estimate our initial guess  $k_{f_{\text{real}}}$ using the additional RPM spin that L1 provides to our system. 
Specifically, let $\omega_{i}$ now be the RPM for motor $i$ given by the geometric controller without L1 compensation and $\omega_{L1i}$ be the RPM for motor $i$ with L1 compensation. Supposing that L1 adaptation has compensated for the propeller damage, we will have 
\begin{equation}
    k_{f_{\text{model}}}\omega_{i}^2 =  k_{f_{\text{real}}}\omega_{L1i}^2.
    \label{eqn:est_prop_damL1}
\end{equation}
Converting terms we have an initial guess
\begin{equation}
         {k_{f_{\text{real}}}} = k_{f_\text{model}} \cdot \frac{\omega_{i}^2}{\omega_{L1i}^2}.
     \label{est_prop_dam}
\end{equation}

Next, we identify the damaged propellers using our initial estimate $d_i$ within a $5\%$ threshold as
\begin{equation}
        d_i = \begin{cases}   
            0.0 &\quad\text{if ${k_{f_{\text{mis}}}}\leq 0.05$}\\
            k_{f_\text{model}} \cdot \frac{\omega_{i}^2}{\omega_{L1i}^2} &\quad\text{if ${k_{f_{\text{mis}}}}> 0.05$}\\
        \end{cases}.
        \label{init_est}
\end{equation}
This threshold is set as undamaged propellers are likely to have negative damage due to torque mismatch. However a negative damage is physically impossible. A detailed explanation of this concept is discussed in Section~\ref{sec:discussion}.

We then formulate our estimate as a quadratic programming problem. Let Let $\mathbf{k} = \begin{bmatrix}
    k_{f_{1\text{real}}} & k_{f_{2\text{real}}} & k_{f_{3\text{real}}} & k_{f_{4\text{real}}} \\
\end{bmatrix}^\top$ represent the true thrust coefficients we wish to solve for and $\mathbf{d}\in \mathbb{R}^{4\times1}$ is a vector where each component $d_i$ is obtained from eq.~(\ref{init_est}). The goal of our optimization is to find a solution that matches the propeller motor dynamics while also being close to our initial estimate described in eq.~(\ref{init_est}) without torque mismatch as
\begin{equation}\label{eqn:op_problem}
\begin{aligned}
\min_{\mathbf{k}} \quad & (\mathbf{k}-\mathbf{d})^\top (\mathbf{k}-\mathbf{d}),\\
\textrm{s.t.} \quad & \mathbf{A} \mathbf{k}=\mathbf{b}.\\
\end{aligned}
\end{equation}

\begin{figure*}[!t]
    \centering
    \includegraphics[width=1\linewidth, trim=0 00 0 0, clip]{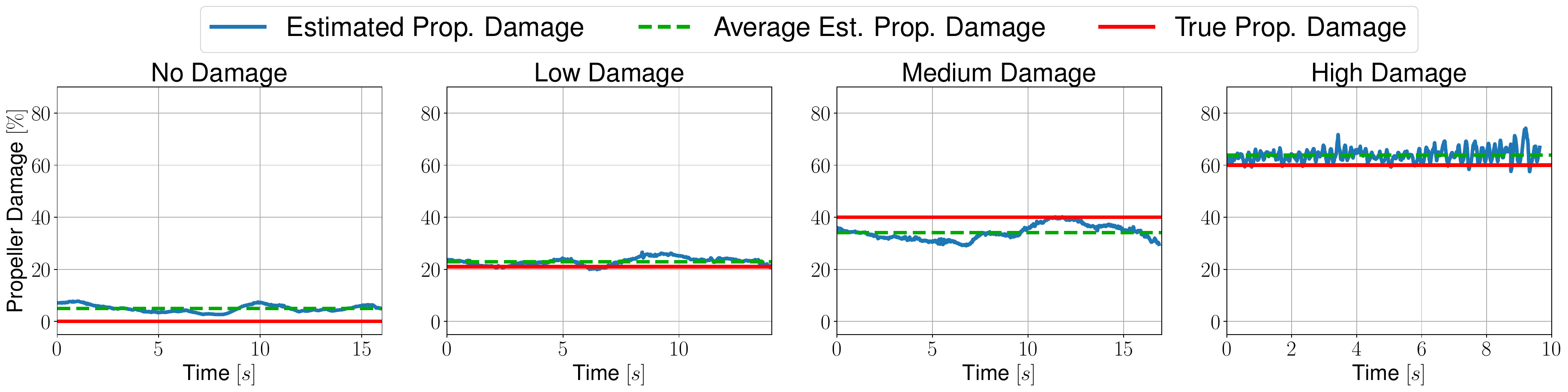}
    \caption{A comparison between estimated damage and true damage with a real-world MAV during a $1.1~\si{m/s}$ flight. Propeller damage is estimated using eq.~(\ref{est_prop_dam}). True Propeller damage is solved by measuring the coefficient on a thrust bench.}
    \label{fig:ratio}
    \vspace{-1.2em}
\end{figure*}

\begin{figure}[!t]
    \centering
    \includegraphics[width=1.0\linewidth, trim=0 00 0 0, clip]{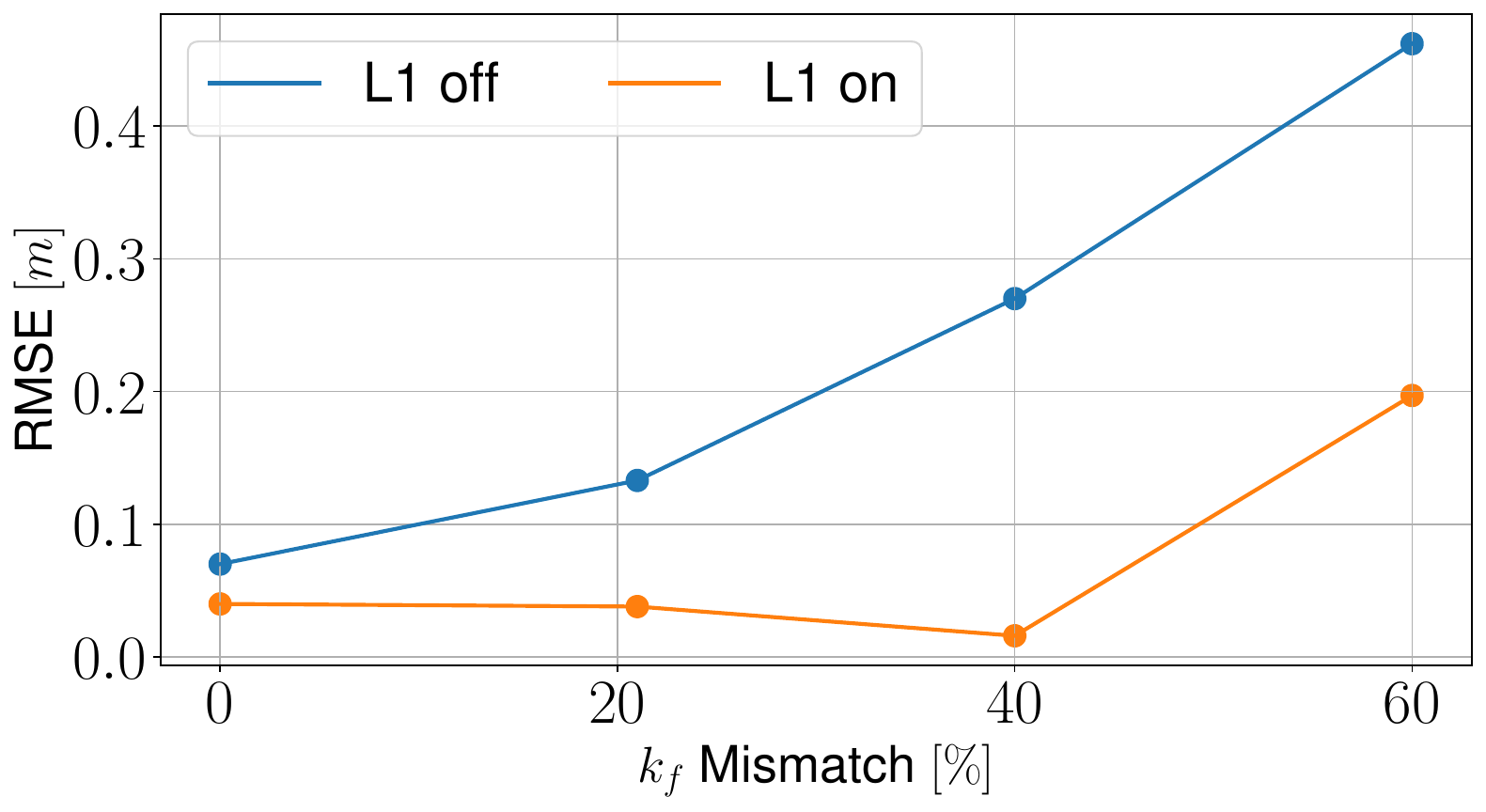}
    \caption{Tracking RMSE averaged over 3 axes vs Propeller Damage at hover with various levels of propeller damage.}
    \label{fig:kf_rmse}
\vspace{-1.3em}
\end{figure}
Here, $\mathbf{A}$ and $\mathbf{b}$ are defined as 
\begin{equation}\label{eq:eq_constr}
\begin{split}
       \mathbf{A}=&\begin{bmatrix}
        \omega_{{L1}_1}^2 & \omega_{{L1}_2}^2&\omega_{{L1}_3}^2 & \omega_{{L1}_4}^2\\
        d_x\omega_{{L1}_1}^2 & d_x\omega_{{L1}_2}^2&-d_x\omega_{{L1}_3}^2 & -d_x\omega_{{L1}_4}^2\\
        -d_y\omega_{{L1}_1}^2 & d_y\omega_{{L1}_2}^2&d_y\omega_{{L1}_3}^2 & -d_y\omega_{{L1}_4}^2\\
    \end{bmatrix},\\
    \mathbf{b}=& \begin{bmatrix}
        f &
        M_1&
        M_2
    \end{bmatrix}^\top,
    \end{split}
\end{equation}
where $\mathbf{b}$ represents the desired  force and moments of the quadrotor commanded by the nominal control without L1 and $\bm{\omega}_{\text{L1}} = \begin{bmatrix} \omega_{{L1}_1} & \omega_{{L1}_2} & \omega_{{L1}_3} & \omega_{{L1}_4} \end{bmatrix}^\top$ are the RPMs from the combination of L1 and nominal geometric system, 
$d_x$ and $d_y$ represent the distance from the propeller to the center of mass in the body frame projected on axes $\mathbf{b}_1$ and $\mathbf{b}_2$ respectively. 
To prevent the torque effect from L1 control we remove the fourth row of our constraints  along with $M_3$ (the torque components) from our optimization.  
The L1 controller drives the error to zero between desired and executed action through the use of supplemental control action\cite{stab_l1}. We construct our equality constraint in eq.~(\ref{eq:eq_constr}) to represent this behaviour where the real thrust coefficients must be consistent with the supplemental action that generates the desired action.
The optimum based on null-space projection is
\begin{equation}\label{eq_closed_form_sol}
    \mathbf{k} = \mathbf{A}^\top\left(\mathbf{A}\mathbf{A}^\top\right)^{-1}\left(\mathbf{b}-\mathbf{A}\mathbf{d}\right) +\mathbf{d}.
\end{equation}
Finally, we convert $\mathbf{k}$ to percentage form using eq.~(\ref{eqn:prop_dam}).

\subsection{Fault-Tolerant Transition}
We leverage our propeller damage estimate, eq.~(\ref{eq_closed_form_sol}) to determine when to switch to fault-tolerant control.
Our fault-tolerant control is implemented based on our previous work \cite{yeom2023geometric} which was only tested in simulation.
From the qualitative results with damaged propellers, we set a threshold at $50\%$ damage. Based on our vehicle's thrust-to-weight ratio of $2.5:1$, we experimentally verify that $50\%$ damage provides a safety margin within the limit.
Once the threshold is met, the system automatically transitions to the fault-tolerant control per our control design as shown in Fig.~\ref{fig:control_diag}.

\begin{figure*}[!t]
    \centering
    \includegraphics[width=1\linewidth, trim=0 00 0 0, clip]{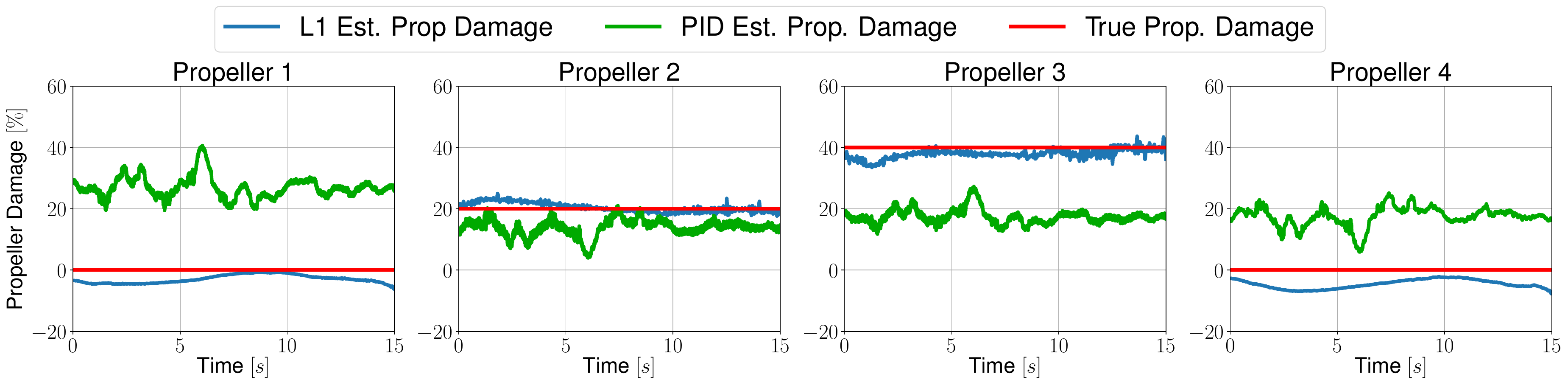}
    \caption{Simultaneous propeller damage (propellers 2 and 3) for MAV moving $0.5\si{m/s}$ in a circle. Both PID (green) and L1 (blue) approaches for propeller damage estimation are shown.}
    \label{fig:multi_prop_dam}
\vspace{-1.2em}
\end{figure*}
\begin{figure*}\centering\includegraphics[width=1\textwidth, trim=0 00 0 0, clip]{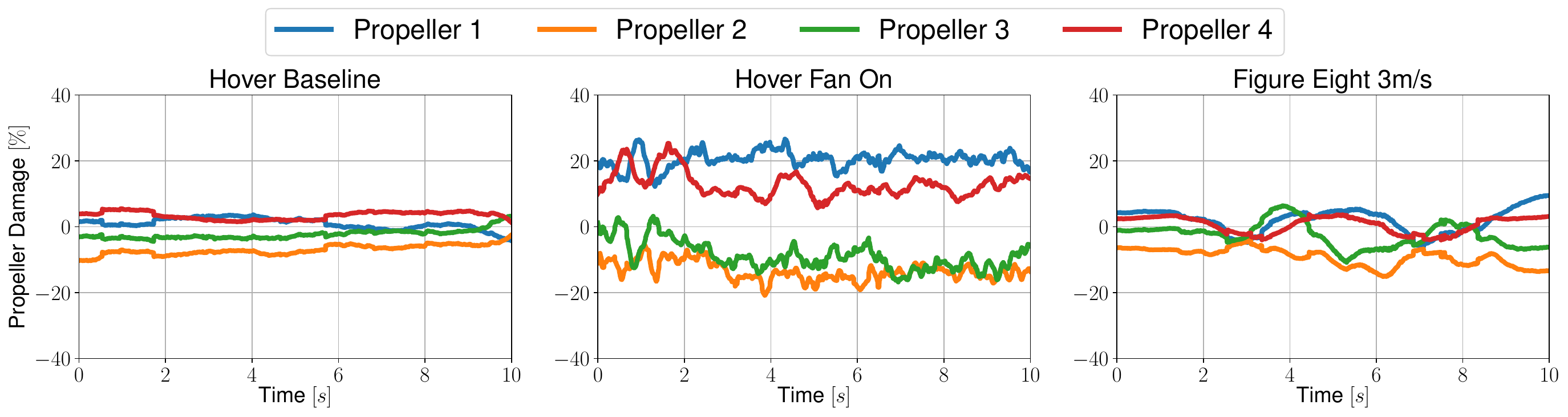}
    \caption{Propeller damage estimate of an undamaged vehicle under 3 different experimental conditions. Hover Baseline refers to a hover with no external forces. Figure Eight refers to flying a trajectory in the shape of an eight with a max velocity of $3~\si{m/s}$. Hover fan on refers to stabilizing the vehicle under a fan blowing wind at $3~\si{m/s}$}
    \vspace{-1.4em}
    \label{fig:extrn_distubr}
\end{figure*}
\section{Results} \label{sec:results}
Our experiments are conducted in a flying space of $10\times6\times4~\si{m^3}$ at the Agile Robotics and Perception Lab (ARPL) at New York University.
The environment is equipped with a Vicon motion capture system which provides accurate pose estimates at $100~\si{Hz}$. This is fused with IMU measurements through an Unscented Kalman Filter to provide state estimates at $500~\si{Hz}$.
The robot, based on our previous work~\cite{LoiannoRAL2017}, is equipped with a VOXL\textsuperscript{\textregistered}2 ModalAI\textsuperscript{TM} and four brushless motors and modified to obtain a $2.5$ to $1$ thrust to weight ratio with total a weight of $700~\si{g}$. 
 We set the $\mathbf{K}$ such that $\lambda = \text{Diag}\begin{bmatrix}
     0.4 & 0.4 & 0.4 & 0.1 & 0.1 & 0.1
 \end{bmatrix}$.

\subsection{L1 Performance Study}
To evaluate the tracking performance, the quadrotor is commanded to follow an ellipse of radius $1$~m in $x$, $0.6$~m in $y$, and $0.1$~m in $z$ for each case of propeller damage as well as under fault-tolerant control with one disabled propeller. The trajectory is completed with and without L1 adaptive control. 
This ellipse is flown at three different speeds: $12$, $8$, and $5 ~\si{s}$ periods. We observe in Table~\ref{tab:rmse1} that the RMSE reduces when the L1 adaptation is switched on in all cases and has lower tracking error than fault-tolerant for propeller damages below $40\%$.
We also see in Fig.~\ref{fig:kf_rmse} the position error grows in the L1 Off case as the severity of propeller damage increases while L1 maintains accurate tracking performance despite the degree of propeller damage. However when the damage reaches greater than $40\%$, the error starts becoming too prominent for the adaptive control to negate. 
In these cases, transitioning to a fault-tolerant controller is preferable. 

\begin{figure*}[h!]
    \centering    \includegraphics[width=1\linewidth]{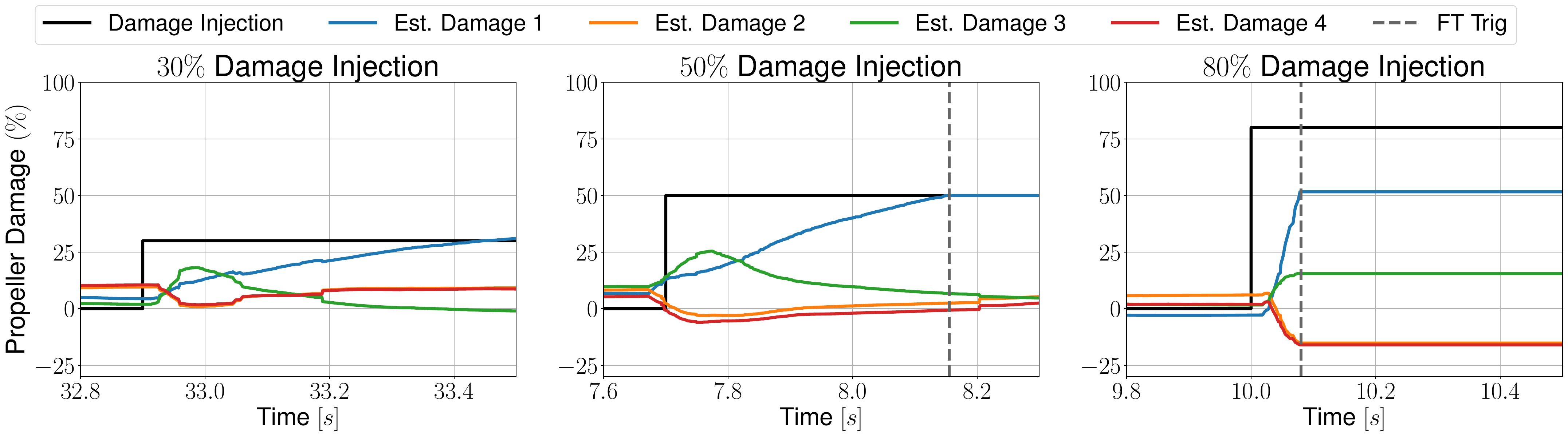}
    \hfill
    \vspace{-1.8em}
    \caption{ Real-world experiments with damage injected to propeller~$1$. Fault-tolerant transition threshold is set to $50\%$.
    }
    \label{fig:damage}
    \vspace{-1.0em}
\end{figure*}
\begin{figure*}[t!]
    \centering
    \includegraphics[width=1\linewidth]{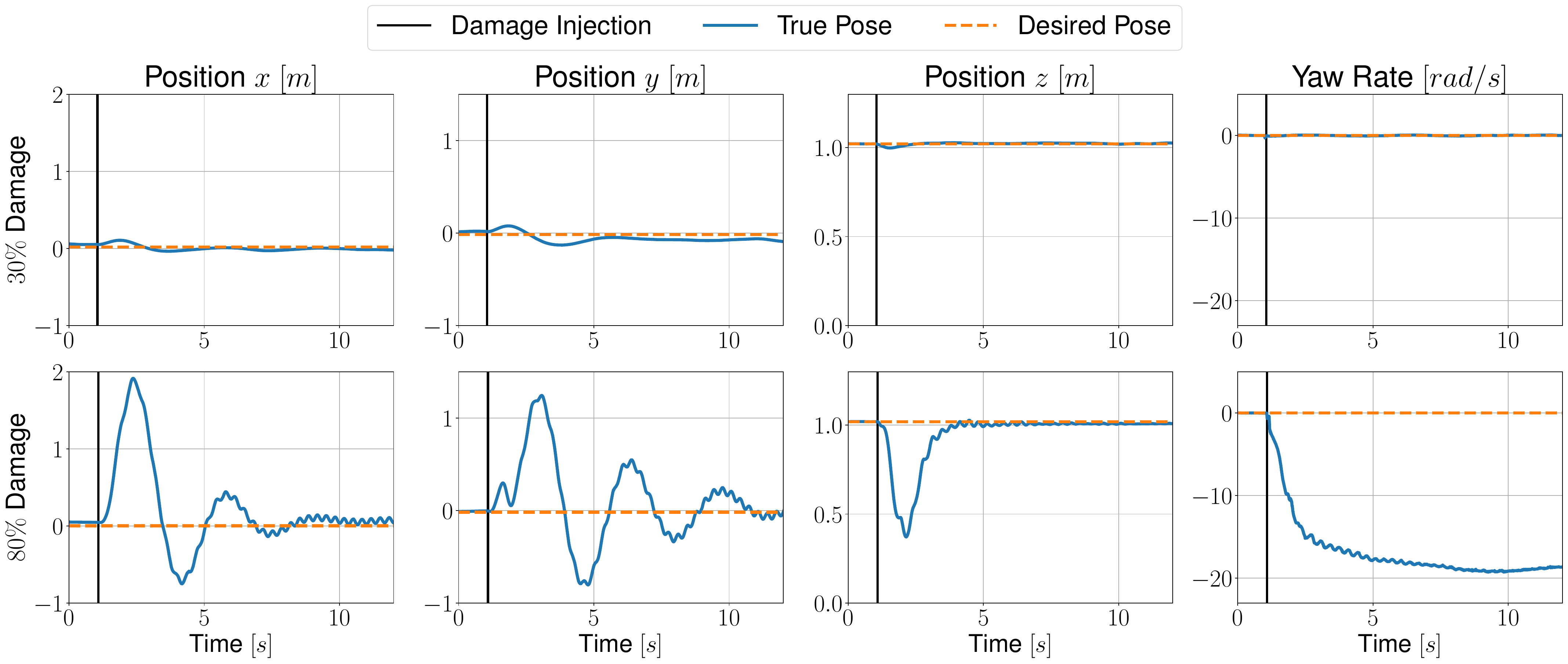}
    \hfill
    \vspace{-1.2em}
    \caption{Position tracking performance after propeller damage is injected in real-world flight. The $80\%$ damage case shows a high yaw rate because it triggers fault-tolerant control, causing a rapid spin unlike the $30\%$ damage case.}
    \label{fig:damage_tracking}
    \vspace{-1.4em}
\end{figure*}
\subsection{Propeller Damage Estimation}
The L1 adaptive controller has demonstrated suitability and robustness to propeller damage compensation.  
In this section, we show that Sec.~\ref{sec:estpropeller_damage}  represents an accurate estimate of the propeller damage, $k_{f_{\text{mis}}}$, even when the aerial robot is in motion. To calculate the various levels of propeller damage, we use a thrust bench to calculate the true thrust coefficient of the damaged propeller, $k_{f_{\text{real}}}$ and undamaged propeller, $k_{f_{\text{model}}}$, coefficient. We then use eq.~(\ref{eqn:prop_dam}) to calculate the damage, $k_{f_{\text{mis}}}$.  We test $4$ levels of propeller damage based on $k_{f_{\text{mis}}}$:
No damage refers to $k_{f_{\text{mis}}}=0\%$. Low damage refers to $k_{f_{\text{mis}}}=20\%$. Medium Damage refers to $k_{f_{\text{mis}}}=40\%$.
High damage refers to $k_{f_{\text{mis}}}=60\%$. In Fig.~\ref{fig:ratio}, we report the estimates of a single damaged propeller at the aforementioned levels while the quadrotor is executing an ellipse trajectory with a max speed of $1.1~\si{m/s}$. This method is able to estimate the error with an error range of $4\%$. In terms of individual actuator forces, this represents $8.3~\si{g}$ of thrust on a $700~\si{g}$ drone. Next, we estimate dual propeller damage simultaneously in Fig.~\ref{fig:multi_prop_dam}.
We compare two methods for damage estimation. First, we use L1 complimentary actions shown in blue. Next, we replace the L1 adaptive control with an integral term in a PID controller for complimentary action in green.
We substitute $\omega_{L1i}$ with the actions generated by an additional integral controller to estimate the propeller damage following the same inference and optimization procedure as Section~\ref{sec:estpropeller_damage}.
The integral term is placed over the velocity expressed in the body frame. 
While the integral term can compensate some damage, it is unsuitable for damage estimation. The L1 complimentary actions on the other hand can be used to accurately estimate the propeller's damage. 

Furthermore, we test the effects of medium speed flight ($3~\si{m/s}$) and external disturbances on our propeller damage estimation. 
In Fig.~\ref{fig:extrn_distubr}, we show the results of 3 experiments. First, we show a control case where the quadrotor is in hover without external disturbances and no damages. Second, we show the same vehicle in hover with $3~\si{m/s}$ wind disturbance. Third, we show a flight without external disturbances flying at $3~\si{m/s}$. Ideally, the desired estimated damage should be $0\%$. Overall, we notice that the fan has a significant effect on damage estimation around $20\%$, but it is not enough to trigger a transition. Medium speed motions at $3 ~\si{m/s}$ instead cause minimal but noticeable deviations from around $0\%$ to $\pm 10\%$.

\subsection{Fault-Tolerant Transition}
We study the rise-time and effectiveness of our transitioning mechanism by artificially injecting a fault mid-flight. This fault is produced by corrupting a motor input to spin slower. 
In  our experiments, we tested the reliability of our system given 40 trials by injecting $30\%$, $40\%$, and $50\%$ damages in a real drone. $30\%$ damage injection has a $100\%$ success rate of not transition whereas the $50\%$ case shows a $100\%$ successful transitions. The $40\%$ injection has a $87.5\%$ success rate for not transitioning and a $12.5\%$ false positive transition. Our method did not produce any false negative.

Additionally, we inject an $80\%$ damage fault in a real-world experiment to show that our latency is low enough for practical uses. The damage estimation of the affected propeller rises to $50\%$ in $0.15~\si{s}$, triggering the transition to fault-tolerant control. 
Fig.~\ref{fig:damage} shows the damage estimation in real-world experiments with $30\%, 50\%,$ and $80\%$ damage injections along with transition times when relevant. When the fault-tolerant controller is activated (i.e., $> 50\%$ damage cases), we stop the inference and estimation process. 
Fig.~\ref{fig:damage_tracking} shows the corresponding position tracking performances for the non transitioning $30\%$ case and the transitioning $80\%$ damage case. We see our methodology can autonomously transition and stabilize the drone despite the injection of heavy propeller damage. For visualization purposes, the last values are repeated instead of cut when the inference stops after the fault tolerant control is triggered. 

\section{Discussion}\label{sec:discussion}
The presented results show the benefit of the proposed adaptive control scheme for damage estimation and compensation. However, L1 adaptation can only correct a disturbance if the actions are within the actuator constraints~\cite{stab_l1}. Therefore a switching threshold for adaptive to fault-tolerant control should reflect this constraint. 
Our system has a $2.5:1$ thrust to weight ratio which enables flight adaptive control with up to $60\%$ damage without switching to fault-tolerant control. We choose to use a switching threshold of $50\%$ to allow some safety margin.
For any quadrotor capable of fault-tolerant flight, a $2:1$ thrust to weight ratio minimum is required. Therefore, a conservative switching threshold of $40\%$ damage can be considered. 
Research has shown that L1 adaptive control is effective on a variety of quadrotors from $27~\si{g}$ Crazyflie~\cite{huang2023datt}, $70~\si{g}$ Parrot Mambo~\cite{NairaL1}, to our heavy weight platform $700~\si{g}$ along with theoretical guarantees \cite{stab_l1}. Our propeller damage estimation method is based on the steady state adaptive properties of L1~\cite{stab_l1}, making it equally generalizable to other platforms.

One limitation of our approach is that we only consider single or dual propeller damages cases. 
Damaged propellers in a near hovering state will spin faster, and undamaged propellers will spin slower or remain at nominal speeds. A reverse effect in hover (lower RPMs of damaged propellers) is highly unlikely, so using the $5\%$ threshold represents a conservative method to detect  propeller damage.
Imagine a single damaged propeller that is spinning clockwise. The damaged propeller must spin faster to match the original thrust. However, this produces less torque than the original undamaged propeller creating a yaw acceleration. 
In order to counteract this yaw acceleration, the two counterclockwise propellers will spin slower. The opposite undamaged clockwise propeller remains consistent with slight deviations. 
The same principle holds for the case of two damaged propellers. 
In the adjacent case, a counterclockwise and clockwise torque both decrease. This results in a cancelling effect where the more damaged propeller's opposite will slow down a constant amount.
In the diagonal case, while a similar thrust can be achieved, the alternative spinning undamaged propellers must reduce their corresponding speeds to counteract the yaw acceleration. These actions are handled by our adaption law eq.~(\ref{eq:gain}), and our optimization method is only for damage estimation. 

\section{Conclusion} \label{sec:conclusion}
In this paper, we presented an adaptive inference and control strategy to ensure safe quadrotor flight in case of propeller damage. Our approach incorporates an L1 adaptive control technique in conjunction with a fault-tolerant mode. 
We propose a method to quantify propeller damage using L1 adaptation that allows smooth transitioning to fault tolerant control in the case of severe damage. The proposed method can effectively account for all range damages from low to severe levels.

Future works will attempt recovery and estimation during more aggressive maneuvers \cite{jeff_robust}. 
We would also like to see if the proposed approach can be extended to other thrust curve models, especially for flight envelopes where the quadratic relation is an inaccurate approximation.
\vspace{-1.0em}
\bibliographystyle{IEEEtran}
\bibliography{main}
\end{document}